\documentclass{article}
\usepackage{arxiv}
\usepackage[utf8]{inputenc}
\usepackage[T1]{fontenc}
\usepackage{hyperref}
\usepackage{url}
\usepackage{booktabs}
\usepackage{amsfonts}
\usepackage{graphicx}
\usepackage{longtable}
\usepackage{array}
\usepackage{calc}
\usepackage{caption}
\usepackage{amsmath}
\usepackage{natbib}
\usepackage{doi}
\usepackage{placeins} 
\newcommand{\hl}[1]{#1} 

\usepackage{authblk}

\title{CedarCypress3D: an annotated UAV-LiDAR dataset of individual trees in
planted cedar and cypress forests}

\author[1]{Katsuto Shimizu\thanks{Corresponding author: shimizu\_katsuto760@ffpri.go.jp}}
\author[2]{Fumiaki Kitahara}
\author[2]{Tomohiro Nishizono}
\author[3]{Hideki Saito}
\author[2]{Masayoshi Takahashi}
\author[4]{Shingo Obata}
\author[4]{Shunsuke Tei}
\author[2]{Naoyuki Furuya}
\author[5]{Tomoya Goto}
\author[2]{Eiji Kodani}
\author[6]{Yusuke Yamada}

\affil[1]{Shikoku Research Center, Forestry and Forest Products Research Institute, \newline 2-915 Asakuranishi, Kochi, Kochi 780-8077, Japan}
\affil[2]{Department of Forest Management, Forestry and Forest Products Research Institute, \newline 1 Matsunosato, Tsukuba, Ibaraki 305-8687, Japan}
\affil[3]{Forestry and Forest Products Research Institute, 1 Matsunosato, Tsukuba, Ibaraki 305-8687, Japan}
\affil[4]{Hokkaido Research Center, Forestry and Forest Products Research Institute, \newline 7 Hitsujigaoka Toyohira, Sapporo, Hokkaido 062-8516, Japan}
\affil[5]{Green Kogyo, Nibancho 5-5, Chiyoda-ku, Tokyo 102-0084, Japan}
\affil[6]{Graduate School of Bioagricultural Sciences, Nagoya University, Furo-cho, \newline Chikusa-ku, Nagoya, 464-8601, Japan}

\renewcommand{\shorttitle}{CedarCypress3D}
\hypersetup{
  pdftitle={CedarCypress3D: an annotated UAV-LiDAR dataset of individual trees in planted cedar and cypress forests},
  pdfauthor={Katsuto Shimizu, Fumiaki Kitahara, Tomohiro Nishizono, Hideki Saito, Masayoshi Takahashi, Shingo Obata, Shunsuke Tei, Naoyuki Furuya, Tomoya Goto, Eiji Kodani, Yusuke Yamada},
  pdfkeywords={Japanese cedar, Japanese cypress, UAV, instance segmentation},
  colorlinks=true, linkcolor=blue, citecolor=blue, urlcolor=blue,
}

\makeatletter
\renewcommand{\@maketitle}{
  \vbox{%
    \hsize\textwidth
    \linewidth\hsize
    \vskip 0.1in
    \@toptitlebar
    \centering
    {\LARGE \@title\par}
    \@bottomtitlebar
    \def\And{%
      \end{tabular}\hfil\linebreak[0]\hfil%
      \begin{tabular}[t]{c}\bf\rule{\z@}{24\p@}\ignorespaces%
    }
    \def\AND{%
      \end{tabular}\hfil\linebreak[4]\hfil%
      \begin{tabular}[t]{c}\bf\rule{\z@}{24\p@}\ignorespaces%
    }
    \begin{tabular}[t]{c}\bf\rule{\z@}{24\p@}\@author\end{tabular}%
    \vskip 0.2in
  }
}
\makeatother

\begin{document}
\maketitle

\begin{abstract}
Individual tree measurements derived from Light Detection and Ranging (LiDAR)
mounted on Unmanned Aerial Vehicles (UAV) provide valuable information for
forest inventory, ecosystem monitoring, and sustainable forest management.
Recent advancements in machine learning have increased the demand for
annotated datasets to develop and evaluate point cloud-based approaches,
especially for individual tree segmentation. However, publicly available
annotated UAV-LiDAR datasets in temperate forests are limited. In this
article, we present \textbf{CedarCypress3D}, a manually annotated UAV-LiDAR
dataset collected in Japanese cedar (\emph{Cryptomeria japonica}) and
Japanese cypress (\emph{Chamaecyparis obtusa}) plantations in Japan. The
dataset consists of UAV-LiDAR point clouds and field survey measurements from
34 circular plots across two sites with different topographic
characteristics, along with terrestrial LiDAR point clouds available for a
subset of 22 plots. A total of 1,627 trees were measured in the census field
survey and manually annotated to match the corresponding trees in the
UAV-LiDAR point clouds. For the subset of plots with terrestrial LiDAR data,
semantic labels (i.e., stem and non-stem) were additionally assigned to tree
points in the UAV-LiDAR data. CedarCypress3D provides high-quality annotated
UAV-LiDAR data for developing and evaluating individual tree instance
segmentation and semantic segmentation methods in temperate planted forests.
The dataset can also support research on tree attribute prediction and
multi-platform LiDAR analysis. The dataset is publicly available at
\url{https://doi.org/10.5281/zenodo.22168721}.
\end{abstract}


\FloatBarrier
\section{Introduction}\label{introduction}

Quantifying individual tree attributes is fundamental to sustainable forest management as it supports carbon accounting and silvicultural decision-making. Traditionally, tree attributes have been collected through field surveys. Although field measurements provide accurate information, they are labor-intensive and time-consuming, leading to limited spatial coverage. Recent advances in Light Detection and Ranging (LiDAR) remote sensing have improved the acquisition of three-dimensional forest structural data \citep{balestraLiDARDataFusion2024,feiglCloseRangeRemoteSensing2025}. LiDAR point clouds capture detailed information on tree crowns, stems, and spatial arrangement, enabling a better understanding of forest structure \citep{liangCloseRangeRemoteSensing2022}. These three-dimensional data enable the development of algorithms for individual tree detection, segmentation, and tree attribute estimation for forest inventories and ecological studies \citep{whiteEnhancedForestInventories2025,takeshigeHighresolutionDigitalCanopy2025,erfanifardContributionHighresolutionRemote2025}.

The emergence of Unmanned Aerial Vehicle (UAV) LiDAR has expanded the applications of LiDAR in forestry. Compared to conventional airborne LiDAR campaigns, UAV-LiDAR can be more easily operated over relatively small forest stands with dense point clouds often exceeding 100 points/m\textsuperscript{2} \citep{liangCloseRangeRemoteSensing2022}. The dense point clouds enable the prediction of detailed individual tree attributes, such as diameter at breast height (DBH), crown base height, and crown structural metrics \citep{htooDevelopmentCrownbasedAllometric2025,eisenschinkForestVariablesLiDAR2025,yangRevealingSpatialDistribution2025,fuComparisonUAVLiDARdrivenBiomass2025}. Recent advances in deep learning have further improved individual tree analysis of LiDAR point clouds, especially for instance segmentation and tree attribute prediction \citep{strakerInstanceSegmentationIndividual2023,luPointCloudUnderstanding2025,luoInceptionFormerDeepLearning2026}. These approaches typically require large and accurately annotated point cloud datasets for model training and evaluation. However, publicly available annotated datasets suitable for individual tree analysis remain limited, particularly for specific forest types and regions \citep{ardohainDeepLearningApplications2026}.

In recent years, the number of publicly available annotated UAV-LiDAR point cloud datasets in forestry has been increasing. The FOR-instance dataset \citep{pulitiFORinstanceUAVLaser2023} is one of the most widely used benchmarks \citep{strakerInstanceSegmentationIndividual2023,xiangAutomatedForestInventory2024,xiangForestFormer3DUnifiedFramework2025,luPointCloudUnderstanding2025}. Other publicly available datasets, such as ForestScan \citep{chavana-bryantForestScanUniqueMultiscale2026}, Lin3D \citep{luPointCloudUnderstanding2025}, and that by Weiser et al. (2022), contain UAV-LiDAR data, in some cases together with terrestrial laser scanning (TLS) and mobile laser scanning (MLS) data collected in tropical and temperate forests \citep{weiserIndividualTreePoint2022,baiSemanticSegmentationSparse2023,luPointCloudUnderstanding2025,chavana-bryantForestScanUniqueMultiscale2026}. However, the geographic and forest-type coverage of existing datasets remains limited, with relatively few datasets available for temperate planted forests. In Japan, planted forests predominantly consist of Japanese cedar (\emph{Cryptomeria japonica}) and Japanese cypress (\emph{Chamaecyparis obtusa}) on steep terrain \citep{takahashiEnhancingThinningEfficiency2026}. Although these species typically have straight trunks, planted forests are often characterized by high stand densities, closed canopies, and abundant understory vegetation. These structural characteristics increase crown overlap and occlusion, which makes individual tree segmentation more challenging than in many existing datasets. Consequently, models trained on datasets from other forest types or geographic regions may not perform well in Japanese planted forests. This highlights the need for a high-quality, annotated UAV-LiDAR dataset that specifically represents temperate planted forests.

In this article, we present \textbf{CedarCypress3D}, an annotated UAV-LiDAR point cloud dataset from temperate planted forests in Japan. The dataset contains UAV-LiDAR point clouds collected in Japanese cedar and Japanese cypress forests, together with census field survey plot data. Terrestrial LiDAR point clouds are also provided for selected survey plots. Both the UAV-LiDAR and terrestrial LiDAR point clouds were manually annotated at the individual-tree level (instance segmentation), and for the subset of plots with corresponding terrestrial LiDAR data, points within individual trees in the UAV-LiDAR point clouds were additionally assigned semantic labels (stem and non-stem). By providing ready-to-use annotated point clouds and corresponding field measurements, CedarCypress3D facilitates the development, benchmarking, and evaluation of algorithms for individual tree analysis, such as instance segmentation, semantic segmentation, tree detection, and tree attribute prediction.

\FloatBarrier
\section{Materials and Methods}\label{materials-and-methods}

\subsection{Study area}\label{study-area}

CedarCypress3D consists of UAV-LiDAR, terrestrial LiDAR, and field survey data acquired at two sites in Oita prefecture, western Japan (Figure 1). A total of 34 circular survey plots were established across the two sites. The first site is located in planted forests in the Saiki region (32.81$^\circ$N, 131.66$^\circ$E), and the second site is located in the Kokonoe region (33.24$^\circ$N, 131.20$^\circ$E). At both sites, the dominant planted conifer species are Japanese cedar (\emph{Cryptomeria japonica}) and Japanese cypress (\emph{Chamaecyparis obtusa}). Stand ages ranged from 25 to 85 years at the Saiki and from 12 to 63 years at the Kokonoe site \citep{forestryagencyNationalForestResource2026}. The Saiki site is characterized by steep terrain, with a mean topographic slope of 28.8$^\circ$ (12.0--37.6$^\circ$), averaged across the survey plots and a mean elevation of approximately 270 m above sea level. In contrast, the Kokonoe site has relatively gentle terrain, with a mean topographic slope of 13.0$^\circ$ (6.5--24.1$^\circ$) and a mean elevation of approximately 800 m. The mean annual temperature is 16.7 $^\circ$C at both sites.

\begin{figure}[htbp]
  \centering
  \includegraphics[width=0.85\linewidth]{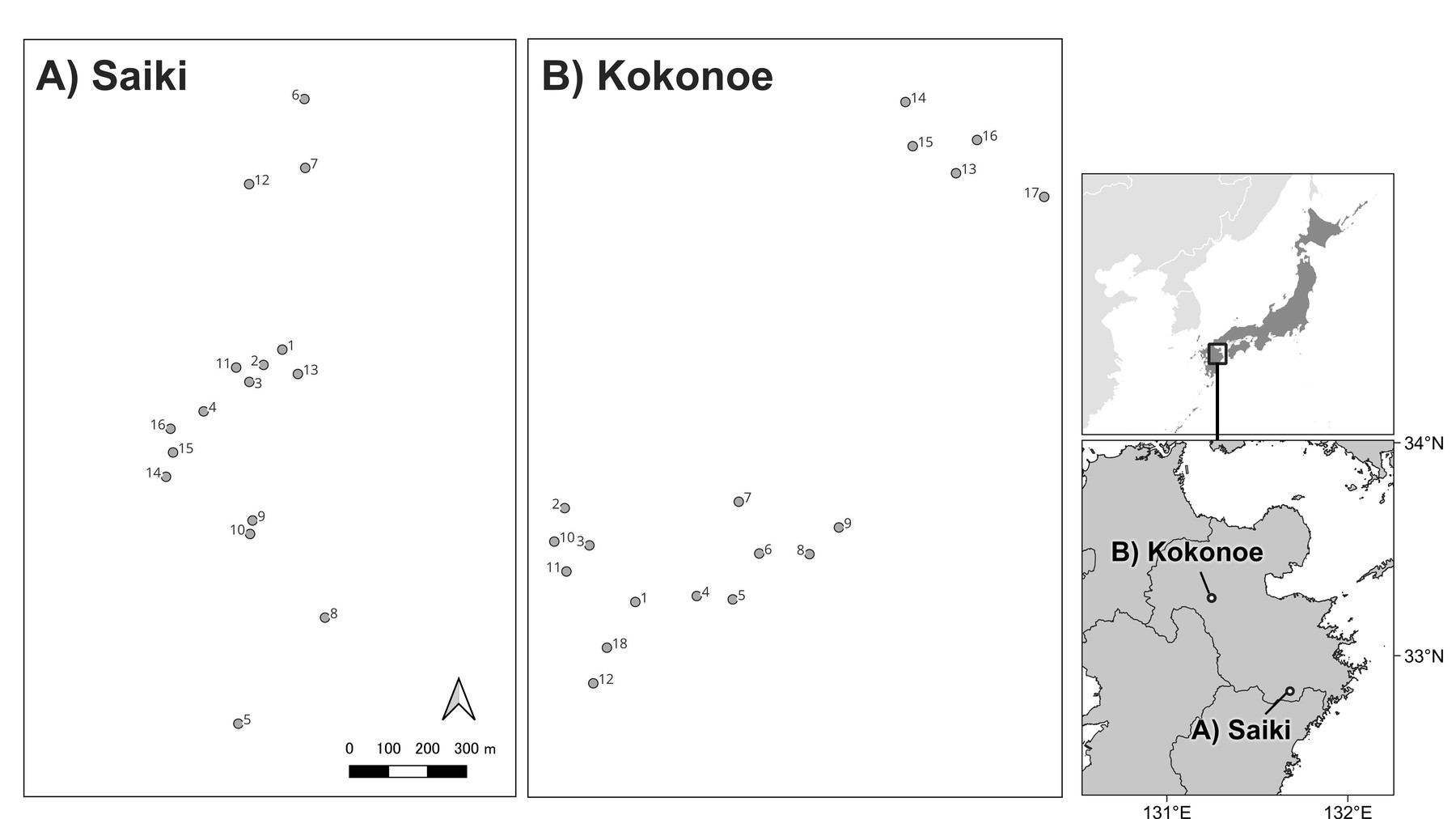}
  \caption{Locations of the survey plots with a radius of 12 m. Latitude and longitude labels have been removed from the main maps. The prefectural boundary data from the National Land Numerical Information were used.}
  \label{fig:1}
\end{figure}

\FloatBarrier

\subsection{Field survey}\label{field-survey}

A field survey was conducted at the Saiki site in January and April 2024. Circular plots with a radius of 12 m (0.04524 ha) were established. All planted trees within each plot were measured regardless of DBH, and other trees with DBH $\geq$ 10 cm were also measured. Of these survey plots, 16 were located in planted forests and included in the dataset. DBH was measured 1.2 m above the ground using a diameter tape. Tree and canopy base heights were measured using a Vertex 5 hypsometer (Haglöf, Sweden). In this study, canopy base height was defined as the vertical distance from the ground to the base of the lowest continuous live branch. Therefore, canopy base height was not recorded for dead trees. Because tree height measurement for one tree was omitted during the field survey, it was imputed using the DBH--height relationship of the measured trees within the same plot based on the Näslund function. Tree form classes were assigned to each tree \emph{in situ} following the Terazaki's tree form classification (Table 1) \citep{terazakiNotesDevelopmentApplication1951}. Tree locations relative to the plot center were measured using a TruPulse 360 laser rangefinder (Laser Technology Inc, USA). The coordinates of each plot center were recorded using a Geode GNS2 Global Navigation Satellite System (GNSS) receiver (Juniper Systems, USA). Individual tree stem volumes were calculated from tree height, DBH, and species, based on stem volume equations \citep{japanforestryagencyTreeVolumeTable1970,japanforestryagencyTreeVolumeTable1970a,hosodaDifferencesPresentStem2010} using the \emph{stemv} package in R \citep{shimizuStemvPackageCalculating2024}. The mean stand volume in Saiki was 952.4 m\textsuperscript{3}/ha in the Japanese cedar plots and 547.4 m\textsuperscript{3}/ha in the Japanese cypress plots. Aboveground biomass (AGB) was also calculated from stem volume using published biomass conversion equations \citep{greenhousegasinventoryofficeofjapanNationalGreenhouseGas2021,shimizuMappingThreeDecades2026}.

A field survey was conducted at the Kokonoe site in April 2025. A total of 18 circular plots were established and measured. The same protocol used at the Saiki site was applied to collect individual tree attributes (DBH, tree height, canopy base height, tree form class, and tree species), tree locations, and plot center coordinates. Geode GNS2 and GNS3 GNSS receivers were used to determine the plot center coordinates at the Kokonoe site. The mean stand volume was 853.4 m\textsuperscript{3}/ha in the Japanese cedar plots and 539.7 m\textsuperscript{3}/ha in the Japanese cypress plots. Across the Saiki and Kokonoe sites, a total of 1,627 trees were measured in 34 plots. Although these plots were dominated by either Japanese cedar or Japanese cypress, a small number of other tree species were also observed. The dataset included 571 Japanese cedar trees, 1,051 Japanese cypress trees, and 5 trees of other species. Summary statistics of the forest structural attributes are presented in Table 2.

\newpage

Table 1. Description of Terazaki's tree form classification used in the field survey.

{\def\LTcaptype{none} 
\begin{longtable}[]{@{}
  >{\centering\arraybackslash}p{(\linewidth - 4\tabcolsep) * \real{0.1389}}
  >{\raggedright\arraybackslash}p{(\linewidth - 4\tabcolsep) * \real{0.1401}}
  >{\raggedright\arraybackslash}p{(\linewidth - 4\tabcolsep) * \real{0.7210}}@{}}
\toprule\noalign{}
\begin{minipage}[b]{\linewidth}\centering
\textbf{Tree form class}
\end{minipage} & \begin{minipage}[b]{\linewidth}\centering
\textbf{Group}
\end{minipage} & \begin{minipage}[b]{\linewidth}\centering
\textbf{Description}
\end{minipage} \\
\midrule\noalign{}
\endhead
\bottomrule\noalign{}
\endlastfoot
1 & Dominant & Tree occupying sufficient growing space for crown development, with crown growth not suppressed by neighboring trees and showing no defects in stem/crown form. \\
2a & Dominant & Tree with an overgrown crown that is horizontally spreading or markedly flat in the upper part of the crown. \\
2b & Dominant & Tree with a poorly developed crown and a smaller stem diameter. \\
2c & Dominant & Tree located between neighboring trees, with insufficient space for crown development, resulting in a one-sided crown. \\
2d & Dominant & Tree with stem defects, such as a forked or swept stem. \\
2e & Dominant & Tree with a damaged stem or crown. \\
3 & Suppressed & Tree with delayed growth, but whose crown is not yet totally suppressed and retains the potential for future height growth. \\
4 & Suppressed & Tree whose crown is already suppressed, but remains alive. \\
5 & Suppressed & Dead tree. \\
\end{longtable}
}

Table 2. Summary statistics of field survey measurements for main tree species at Saiki and Kokonoe sites. Values in parentheses indicate the minimum and maximum values. DBH, diameter at breast height; TH, tree height; BA, basal area. Main species indicates the dominant species in each plot; the stand statistics therefore refer to all trees within those plots, not only to individuals of the dominant species.

{\def\LTcaptype{none} 
\begin{longtable}[]{@{}
  >{\centering\arraybackslash}p{(\linewidth - 14\tabcolsep) * \real{0.1067}}
  >{\centering\arraybackslash}p{(\linewidth - 14\tabcolsep) * \real{0.1083}}
  >{\raggedleft\arraybackslash}p{(\linewidth - 14\tabcolsep) * \real{0.0807}}
  >{\raggedright\arraybackslash}p{(\linewidth - 14\tabcolsep) * \real{0.1384}}
  >{\raggedright\arraybackslash}p{(\linewidth - 14\tabcolsep) * \real{0.1264}}
  >{\raggedright\arraybackslash}p{(\linewidth - 14\tabcolsep) * \real{0.1264}}
  >{\raggedright\arraybackslash}p{(\linewidth - 14\tabcolsep) * \real{0.1745}}
  >{\raggedright\arraybackslash}p{(\linewidth - 14\tabcolsep) * \real{0.1384}}@{}}
\toprule\noalign{}
\begin{minipage}[b]{\linewidth}\centering
\textbf{Site}
\end{minipage} & \begin{minipage}[b]{\linewidth}\centering
\textbf{Main species}
\end{minipage} & \begin{minipage}[b]{\linewidth}\centering
\textbf{n plots}
\end{minipage} & \begin{minipage}[b]{\linewidth}\centering
\textbf{Ave. trees /ha}
\end{minipage} & \begin{minipage}[b]{\linewidth}\centering
\textbf{Ave. DBH}

\textbf{(cm)}
\end{minipage} & \begin{minipage}[b]{\linewidth}\centering
\textbf{Ave. TH (m)}
\end{minipage} & \begin{minipage}[b]{\linewidth}\centering
\textbf{Ave. stand volume (m\textsuperscript{3}/ha)}
\end{minipage} & \begin{minipage}[b]{\linewidth}\centering
\textbf{Ave. BA (m\textsuperscript{2}/ha)}
\end{minipage} \\
\midrule\noalign{}
\endhead
\bottomrule\noalign{}
\endlastfoot
Saiki & Japanese cedar & 11 & \begin{minipage}[t]{\linewidth}\raggedleft
940\\
(420-1592)\strut
\end{minipage} & \begin{minipage}[t]{\linewidth}\raggedleft
35.1\\
(12.8-76.6)\strut
\end{minipage} & \begin{minipage}[t]{\linewidth}\raggedleft
24.0\\
(12-41.1)\strut
\end{minipage} & \begin{minipage}[t]{\linewidth}\raggedleft
952.4\\
(396-2012.9)\strut
\end{minipage} & \begin{minipage}[t]{\linewidth}\raggedleft
86.5\\
(47.4-156.2)\strut
\end{minipage} \\
& Japanese cypress & 5 & \begin{minipage}[t]{\linewidth}\raggedleft
1561\\
(752-2454)\strut
\end{minipage} & \begin{minipage}[t]{\linewidth}\raggedleft
22.6\\
(10-43)\strut
\end{minipage} & \begin{minipage}[t]{\linewidth}\raggedleft
17.5\\
(5.2-25.6)\strut
\end{minipage} & \begin{minipage}[t]{\linewidth}\raggedleft
547.4\\
(235.8-715.3)\strut
\end{minipage} & \begin{minipage}[t]{\linewidth}\raggedleft
59.7\\
(37.1-81.3)\strut
\end{minipage} \\
Kokonoe & Japanese cedar & 2 & \begin{minipage}[t]{\linewidth}\raggedleft
1304\\
(1105-1503)\strut
\end{minipage} & \begin{minipage}[t]{\linewidth}\raggedleft
28.9\\
(11-42.3)\strut
\end{minipage} & \begin{minipage}[t]{\linewidth}\raggedleft
20.9\\
(13.3-26.5)\strut
\end{minipage} & \begin{minipage}[t]{\linewidth}\raggedleft
853.4\\
(631.5-1075.4)\strut
\end{minipage} & \begin{minipage}[t]{\linewidth}\raggedleft
86.1\\
(77.3-95.0)\strut
\end{minipage} \\
  & Japanese cypress & 16 & \begin{minipage}[t]{\linewidth}\raggedleft
951\\
(619-1614)\strut
\end{minipage} & \begin{minipage}[t]{\linewidth}\raggedleft
29.4\\
(9.2-74.2)\strut
\end{minipage} & \begin{minipage}[t]{\linewidth}\raggedleft
17.3\\
(6.8-25.9)\strut
\end{minipage} & \begin{minipage}[t]{\linewidth}\raggedleft
539.7\\
(293.3-896.4)\strut
\end{minipage} & \begin{minipage}[t]{\linewidth}\raggedleft
63.6\\
(45.5-88.2)\strut
\end{minipage} \\
\end{longtable}
}

\subsection{UAV-LiDAR measurements}\label{uav-lidar-measurements}

Different UAV-LiDAR systems were used at Saiki and Kokonoe. At the Saiki site, a RIEGL VUX-1LR sensor mounted on a FAZER R G2 UAV was used to acquire the LiDAR point cloud data (Table 3). The flight campaign was conducted between April 12 and 18, 2022, at a flight altitude of approximately 80--130 m above ground level. The point cloud data had an average point density of 5,802 points/m\textsuperscript{2} across all 16 plots at the site (Table 4). At the Kokonoe site, a DJI Zenmuse L2 sensor mounted on a Matrice 300 RTK UAV was used to acquire LiDAR point cloud data between November 18 and 20, 2024. The flight altitude was approximately 100 m above ground level, with an average point density of 676 points/m\textsuperscript{2}. The UAV-LiDAR and field measurements were separated by two growing seasons at the Saiki site, whereas they were conducted within the same growing season at the Kokonoe site.

Both sets of UAV-LiDAR data were processed using the \emph{lidR} package \citep{rousselLidRPackageAnalysis2020}. First, an isolated voxel filter was applied to remove outlier points. For the Saiki data, the progressive morphological filter algorithm was used to classify ground points because no ground classification was provided by the data vendor. In contrast, the existing ground classification provided by the data vendor was used for the Kokonoe data.

\newpage

Table 3. Specifications of the LiDAR sensors used in the dataset. FWHM, Full Width at Half Maximum.

{\def\LTcaptype{none} 
\begin{longtable}[]{@{}
  >{\centering\arraybackslash}p{(\linewidth - 6\tabcolsep) * \real{0.2879}}
  >{\raggedright\arraybackslash}p{(\linewidth - 6\tabcolsep) * \real{0.2277}}
  >{\raggedright\arraybackslash}p{(\linewidth - 6\tabcolsep) * \real{0.2277}}
  >{\raggedright\arraybackslash}p{(\linewidth - 6\tabcolsep) * \real{0.2277}}@{}}
\toprule\noalign{}
\begin{minipage}[b]{\linewidth}\centering
\textbf{Specification}
\end{minipage} & \begin{minipage}[b]{\linewidth}\centering
\textbf{VUX-1LR}
\end{minipage} & \begin{minipage}[b]{\linewidth}\centering
\textbf{Zenmuse L2}
\end{minipage} & \begin{minipage}[b]{\linewidth}\centering
\textbf{UTM-30LX-EW}
\end{minipage} \\
\midrule\noalign{}
\endhead
\bottomrule\noalign{}
\endlastfoot
Platform & FAZER R G2 & Matrice 300 RTK & OWL \\
Type & UAV & UAV & Terrestrial \\
Max. pulse repetition rate & 1,500 kHz & 240 kHz & 43.2 kHz \\
Max. measuring range & 1350 m

(@60\% reflectivity) & 450 m

(@50\% reflectivity) & 30 m

(@90\% reflectivity) \\
Ranging accuracy & $\pm$15 mm & $\pm$20 mm (@150 m) & $\pm$50 mm (@30 m) \\
Beam divergence & 0.5 mrad

(1/e\textsuperscript{2}) & 0.2 $\times$ 0.6 mrad

(FWHM) & - \\
Angular resolution & - & - & 4.4 mrad (0.25$^\circ$) \\
\end{longtable}
}

Table 4. Summary of plot measurements in the CedarCypress3D dataset. Main species indicates the dominant species in each plot; the number of trees therefore refers to all trees within those plots, not only to individuals of the dominant species.

{\def\LTcaptype{none} 
\begin{longtable}[]{@{}
  >{\centering\arraybackslash}p{(\linewidth - 14\tabcolsep) * \real{0.1232}}
  >{\centering\arraybackslash}p{(\linewidth - 14\tabcolsep) * \real{0.1946}}
  >{\centering\arraybackslash}p{(\linewidth - 14\tabcolsep) * \real{0.0843}}
  >{\centering\arraybackslash}p{(\linewidth - 14\tabcolsep) * \real{0.1317}}
  >{\centering\arraybackslash}p{(\linewidth - 14\tabcolsep) * \real{0.0972}}
  >{\centering\arraybackslash}p{(\linewidth - 14\tabcolsep) * \real{0.1492}}
  >{\centering\arraybackslash}p{(\linewidth - 14\tabcolsep) * \real{0.1077}}
  >{\centering\arraybackslash}p{(\linewidth - 14\tabcolsep) * \real{0.1121}}@{}}
\toprule\noalign{}
\begin{minipage}[b]{\linewidth}\centering
\textbf{Site}
\end{minipage} & \begin{minipage}[b]{\linewidth}\centering
\textbf{Main species}
\end{minipage} & \begin{minipage}[b]{\linewidth}\centering
\textbf{n plots}
\end{minipage} & \begin{minipage}[b]{\linewidth}\centering
\textbf{n terrestrial LiDAR}
\end{minipage} & \begin{minipage}[b]{\linewidth}\centering
\textbf{n trees}
\end{minipage} & \begin{minipage}[b]{\linewidth}\centering
\textbf{Sensor}
\end{minipage} & \begin{minipage}[b]{\linewidth}\centering
\textbf{Point density (pts/m\textsuperscript{2})}
\end{minipage} & \begin{minipage}[b]{\linewidth}\centering
\textbf{Average slope ($^\circ$)}
\end{minipage} \\
\midrule\noalign{}
\endhead
\bottomrule\noalign{}
\endlastfoot
Saiki & Japanese cedar & 11 & 7 & 468 & RIEGL

VUX-1LR & 5754 & 28.4 \\
& Japanese cypress & 5 & 0 & 353 & RIEGL

VUX-1LR & 5909 & 29.7 \\
Kokonoe & Japanese cedar & 2 & 1 & 118 & DJI

Zenmuse L2 & 694 & 11.0 \\
& Japanese cypress & 16 & 14 & 688 & DJI

Zenmuse L2 & 674 & 13.3 \\
\end{longtable}
}

\subsection{Terrestrial LiDAR measurements}\label{terrestrial-lidar-measurements}

For 22 of the 34 plots (7 in Saiki and 15 in Kokonoe), terrestrial LiDAR data was also collected using the Optical Woods Ledger (OWL) AME-OL104 (AdIn Research, Japan) during the field survey (Table 4). The OWL system is equipped with a UTM-30LX-EW \hl{} laser sensor (Hokuyo, Japan) on a single-legged instrument. This system employs simultaneous localization and mapping (SLAM) and does not require remaining stationary during scanning \citep{tsubouchiIntroductionSimultaneousLocalization2019}. In each plot, nine scans were conducted at a distance of approximately 10 m. The multi-scan data acquired with OWL were processed using the OWL manager software (AdIn Research, Japan) to generate point clouds and tree location information. Ground points were classified using the progressive morphological filter algorithm in the \emph{lidR} package \citep{rousselLidRPackageAnalysis2020}. Due to the laser sensor specification, canopy occlusion has been reported in previous studies, especially for taller trees \citep{kitaharaComparisonForestMeasurement2020,nishizonoEffectsStandCondition2020,shimizuIntegratingTerrestrialLaser2022}. However, the point cloud data from this terrestrial LiDAR system were included to complement the UAV-LiDAR data.

\subsection{Co-registration of field survey, UAV-LiDAR, and terrestrial LiDAR data}\label{co-registration-of-field-survey-uav-lidar-and-terrestrial-lidar-data}

The relative tree locations recorded during the field survey were converted to geographic coordinates using the plot center coordinates from GNSS measurements. Because these tree locations had positional errors associated with GNSS measurements, they were spatially aligned with the UAV-LiDAR data prior to manual annotation using treetop locations provided by the data vendors. These treetop locations were used only for the co-registration process and were not used for subsequent analysis. To align the tree locations in the field surveys with the UAV-LiDAR-derived treetops, a hierarchical 2D rigid registration procedure was applied. First, rotation-invariant geometric features were computed from the sorted distances to the eight nearest neighbors for each tree, and candidate matches were established by selecting the five nearest neighbors in the feature space. Second, a coarse global alignment was performed using the Random Sample Consensus (RANSAC) with 8,000 iterations to estimate tentative rotation and translation parameters. Finally, the alignment was fine-tuned using a trimmed Iterative Closest Point (ICP) algorithm, which iteratively optimized the transformation parameters using only the closest 70\% of the pairs to ensure robustness against outliers. After the final transformation parameters were determined, the UAV-LiDAR point clouds were clipped to a radius of 18 m (12 m plot radius + 6 m buffer) centered on each co-registered plot location (Figures 2 and 3).

\begin{figure}[htbp]
  \centering
  \includegraphics[width=0.80\linewidth]{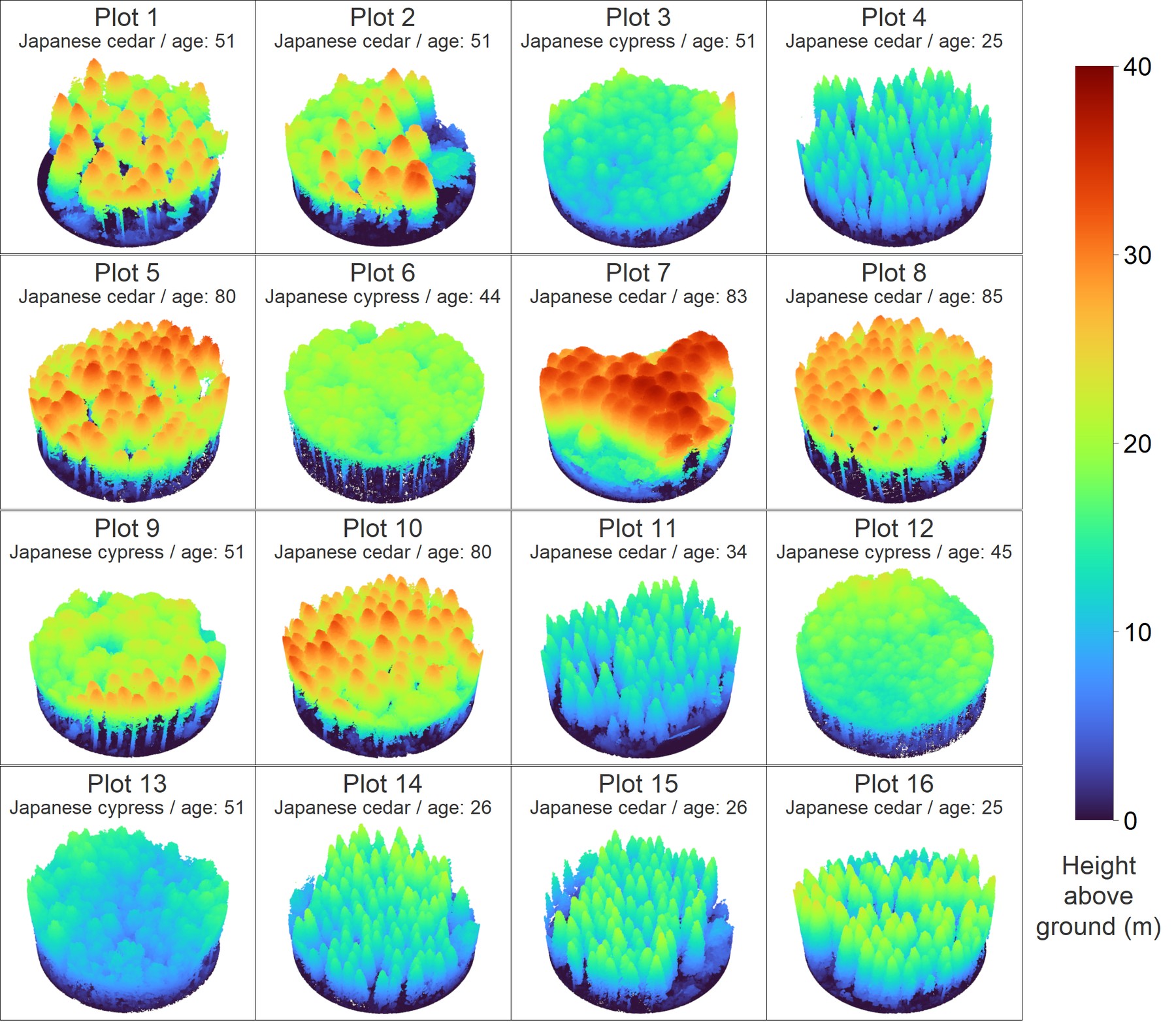}
  \caption{Visualization of the clipped UAV-LiDAR point clouds at the Saiki site.}
  \label{fig:2}
\end{figure}

\begin{figure}[htbp]
  \centering
  \includegraphics[width=0.80\linewidth]{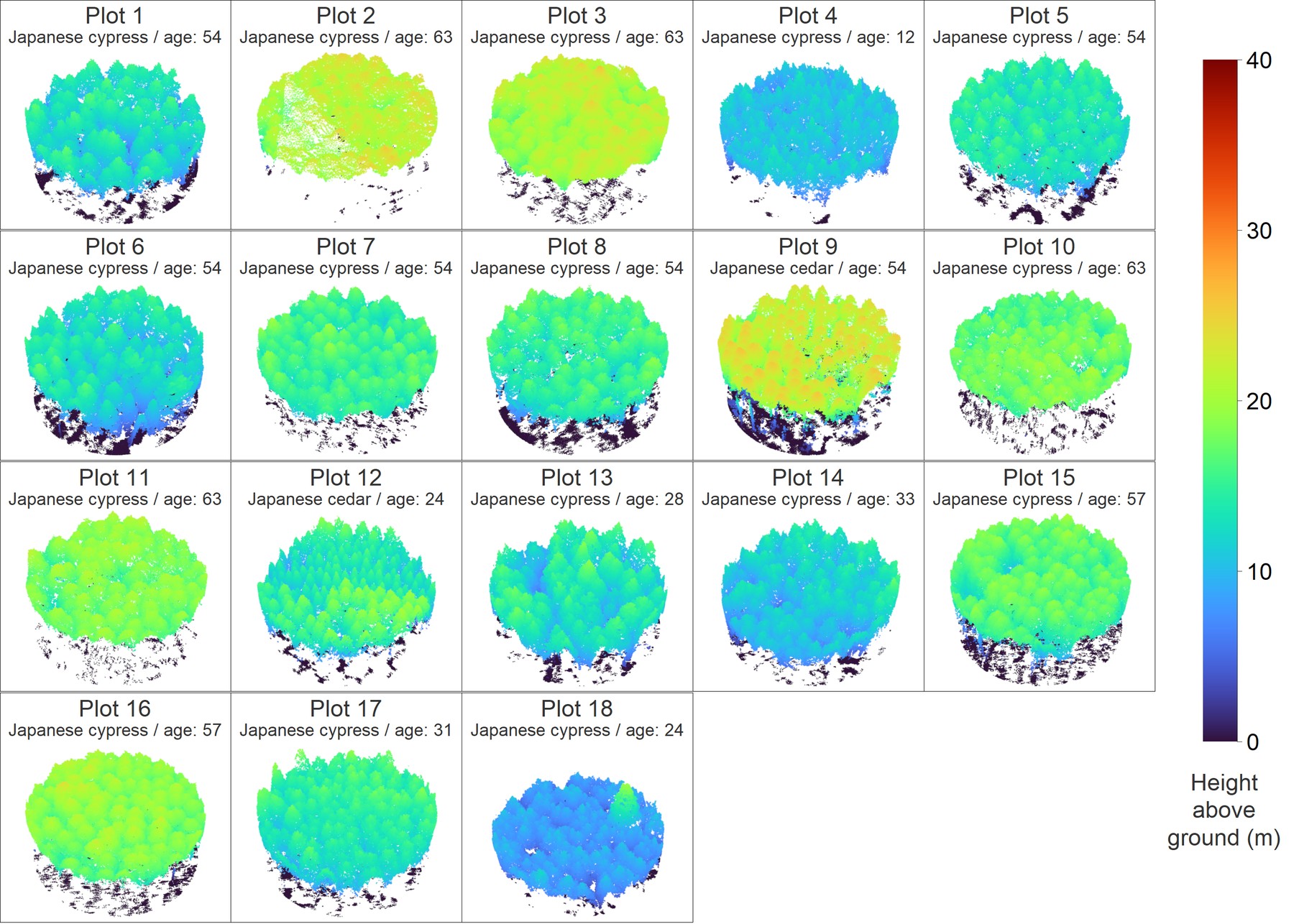}
  \caption{Visualization of the clipped UAV-LiDAR point clouds at the Kokonoe site.}
  \label{fig:3}
\end{figure}

\FloatBarrier

For the OWL terrestrial LiDAR data, a similar co-registration procedure was applied to align the terrestrial LiDAR point clouds with the UAV-LiDAR data. The tree locations derived from terrestrial LiDAR were used as inputs for the same initial coarse registration procedure described above. The estimated transformation parameters were then applied to the original terrestrial LiDAR point clouds. Finally, a fine registration between the terrestrial LiDAR and UAV-LiDAR point clouds was performed using the ICP algorithm. For three plots at the Kokonoe site, manual co-registration was performed instead of ICP, because the ICP failed to align the two point clouds owing to insufficient overlap.

\subsection{Manual annotation}\label{manual-annotation}

The UAV-LiDAR point clouds were manually annotated into individual trees using the segment tool in CloudCompare v2.13.2 (Figure 4). Two annotators performed the initial annotation, which was subsequently reviewed and revised by a third annotator to determine the final labels. All trees within each plot were annotated. To facilitate the annotation process for the Kokonoe data, marker-controlled watershed segmentation was applied to digital canopy height models (DCHM) alongside tree locations recorded during the field survey. This procedure was not applied to the Saiki data. Noise points were also identified during annotation.

After annotation of the UAV-LiDAR point clouds, tree IDs from the field survey were assigned to the annotated trees based on the co-registered tree locations (Figure 5). Each annotated tree was linked to the nearest field-survey tree using the centroid of the annotated point cloud. Because field survey tree locations were co-registered to treetops derived from the UAV-LiDAR data, some trees located near the plot boundary may extend beyond the 12 m plot radius. This discrepancy reflects the positional uncertainty after co-registration and does not indicate that these trees are located outside the plot.

\begin{figure}[htbp]
  \centering
  \includegraphics[width=0.85\linewidth]{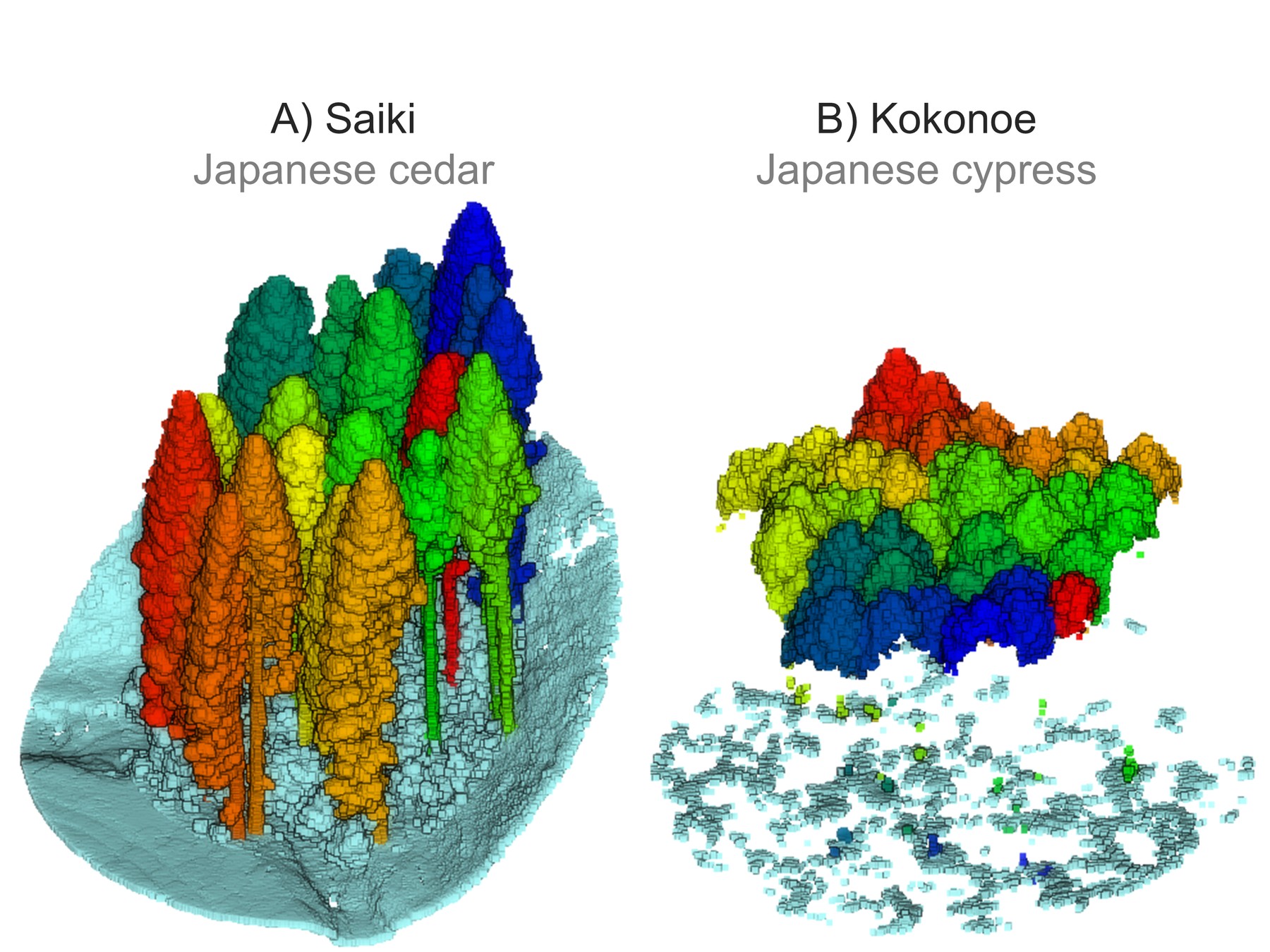}
  \caption{Visualization of individual tree annotations for plots in (A) Saiki (Plot 1) and (B) Kokonoe (Plot 3). Trees and understory vegetation outside the plot boundary are not shown.}
  \label{fig:4}
\end{figure}

\newpage

For the OWL terrestrial LiDAR data, manually annotated UAV-LiDAR data were used to assign initial tree IDs. A k-nearest neighbor (kNN) approach was applied to transfer tree IDs from UAV-LiDAR point clouds to the terrestrial LiDAR point clouds to reduce the time and effort required for manual annotation. For the Saiki data, nearest neighbors were identified using three-dimensional (XYZ) information, whereas only two-dimensional (XY) information was used for the Kokonoe data because the lower UAV-LiDAR point density reduced the reliability of three-dimensional allocation. The automatically assigned tree IDs were then manually reviewed and corrected by a single annotator. While both stem and crown (i.e., branches and leaves) points were corrected for the Saiki data, only minimal correction was applied to crown points for the Kokonoe data, because overlaps between adjacent trees made it difficult to determine which points belonged to the corresponding trees.

For the subset of plots with corresponding terrestrial LiDAR data (n = 22), semantic labels (stem and non-stem) were additionally assigned to points within individual trees in the UAV-LiDAR point clouds. To do this, a single annotator manually classified points as stem or non-stem (i.e., crown) within previously segmented individual trees using CloudCompare. A total of 244 trees from 7 plots at the Saiki site and 624 trees from 15 plots at the Kokonoe site were assigned semantic labels. However, trees at the Kokonoe site frequently lacked sufficient stem points because of lower point density of the UAV-LiDAR data.

\begin{figure}[htbp]
  \centering
  \includegraphics[width=0.82\linewidth]{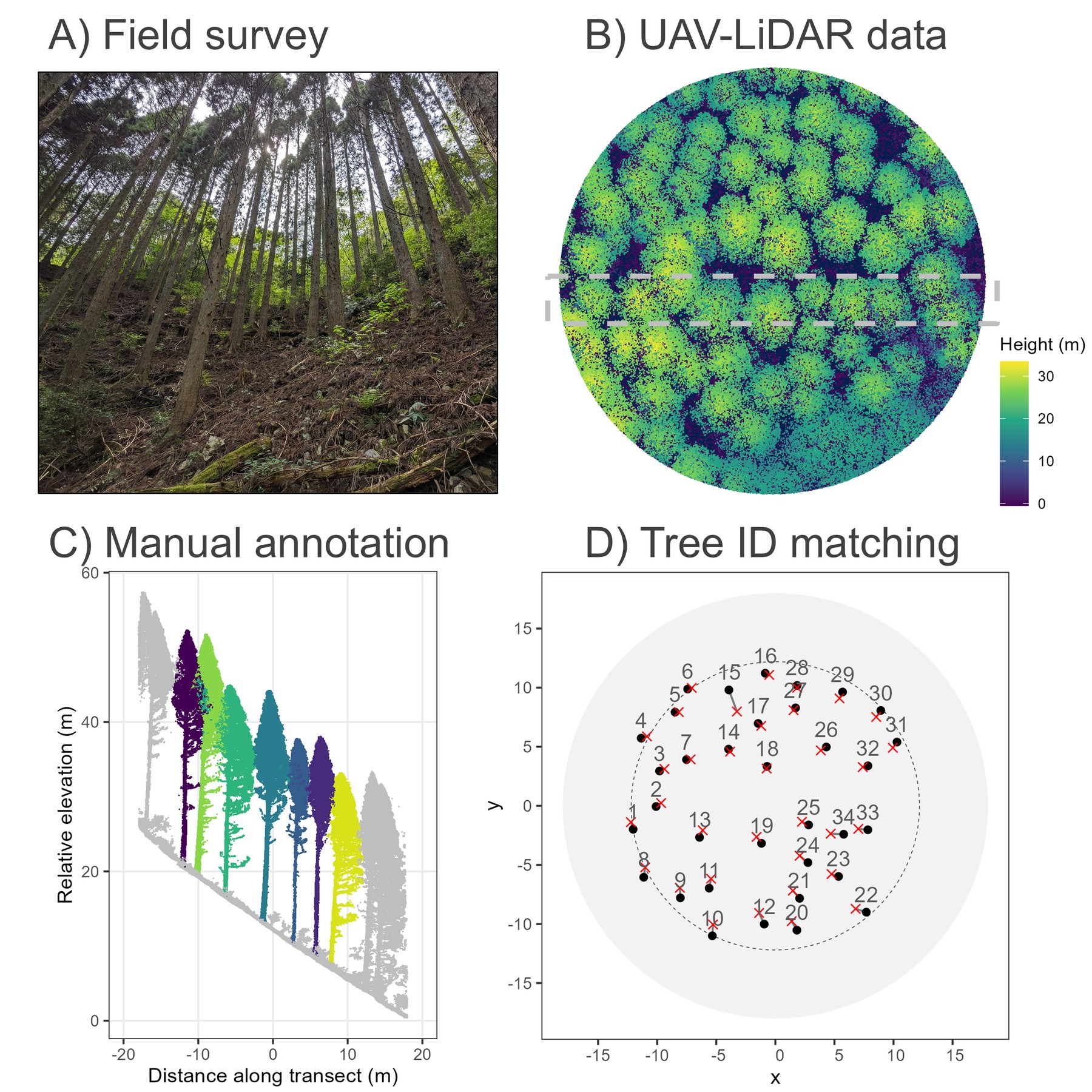}
  \caption{Visualization of tree ID allocation using tree locations recorded during the field survey and annotated UAV-LiDAR point clouds. Examples from Plot 10 at the Saiki site are presented. (A) Photograph of the field survey plot. (B) UAV-LiDAR point cloud colored by normalized height with dashed lines indicating the transect strip shown in (C). (C) Cross-sectional view of the manually annotated point cloud along the transect in (B). (D) Tree ID allocation based on nearest neighbor matching between field-survey tree locations (black) and UAV-LiDAR annotated tree centroids (red).}
  \label{fig:5}
\end{figure}

\FloatBarrier
\section{Dataset description}\label{dataset-description}

The UAV-LiDAR and terrestrial LiDAR point clouds are provided in LAS format. The terrestrial LiDAR data are available for a subset of the plots. The \emph{treeID} attribute represents the annotated tree ID, which corresponds to the tree ID in the field survey data. Non-tree points and trees located outside the plot are assigned \emph{treeID} = 0 (Figure 6). To anonymize the study sites, the point cloud coordinates were translated into a local coordinate system and the original geographic coordinates were removed.

The LAS point classification defined in this study is presented in Table 5. Points outside the plot boundary are assigned to the \textbf{Outside plot} class (Classification = 1). This class includes all points within the buffer around each plot and tree points belonging to trees rooted outside the plot, the crowns of which extended into the plot (Figure 6). \textbf{Ground} points (Classification = 2) were primarily identified using automatic ground classification, with manual corrections applied during annotation. \textbf{Low vegetation} (Classification = 3) includes vegetation within the plot that is neither ground nor tree. Manually annotated \textbf{individual trees within the plot} are assigned Classification = 5. \textbf{Noise} points are assigned Classification = 7.

In addition to the Classification attribute, the \emph{treeComponent} attribute is stored in all UAV-LiDAR LAS data to distinguish stem from non-stem components of individual trees (Figure 6). The attribute is defined as unlabeled (= 0), stem (= 1), or non-stem (= 2) (Table 6). Stem and non-stem labels are available only for tree points (i.e., Classification = 5) in the UAV-LiDAR data from plots with corresponding terrestrial LiDAR measurements. For all other points, including tree points in plots without terrestrial LiDAR measurements, \emph{treeComponent} is set to 0.

Field survey data are provided in CSV format. The \emph{treeID} field corresponds to the \emph{treeID} attribute in the UAV-LiDAR point cloud. Tree attributes measured during the field survey are provided for each tree, including plot ID, tree ID, species, tree form class, DBH, tree height, canopy base height, and stem volume. Plot-level attributes (i.e., survey date and stand age) are also provided.

\bigskip

Table 5. LiDAR point classification used in the dataset.

{\def\LTcaptype{none} 
\begin{longtable}[]{@{}
  >{\centering\arraybackslash}p{(\linewidth - 2\tabcolsep) * \real{0.1791}}
  >{\raggedright\arraybackslash}p{(\linewidth - 2\tabcolsep) * \real{0.4408}}@{}}
\toprule\noalign{}
\begin{minipage}[b]{\linewidth}\centering
\textbf{Classification}
\end{minipage} & \begin{minipage}[b]{\linewidth}\centering
\textbf{Description}
\end{minipage} \\
\midrule\noalign{}
\endhead
\bottomrule\noalign{}
\endlastfoot
1 & Outside plot \\
2 & Ground \\
3 & Low vegetation within the plot \\
5 & Trees within the plot \\
7 & Noise \\
\end{longtable}
}

\bigskip

Table 6. Description of the \emph{treeComponent} attribute in the UAV-LiDAR data. Stem and non-stem classification is available only for plots with corresponding terrestrial LiDAR data.

{\def\LTcaptype{none} 
\begin{longtable}[]{@{}
  >{\centering\arraybackslash}p{(\linewidth - 2\tabcolsep) * \real{0.1395}}
  >{\raggedright\arraybackslash}p{(\linewidth - 2\tabcolsep) * \real{0.7956}}@{}}
\toprule\noalign{}
\begin{minipage}[b]{\linewidth}\centering
\textbf{Value}
\end{minipage} & \begin{minipage}[b]{\linewidth}\centering
\textbf{Description}
\end{minipage} \\
\midrule\noalign{}
\endhead
\bottomrule\noalign{}
\endlastfoot
0 & Unlabeled / no data (points with Classification $\neq$ 5, or tree points in plots without terrestrial LiDAR data) \\
1 & Stem \\
2 & Non-stem (branches and leaves) \\
\end{longtable}
}

\begin{figure}[htbp]
  \centering
  \includegraphics[width=0.85\linewidth]{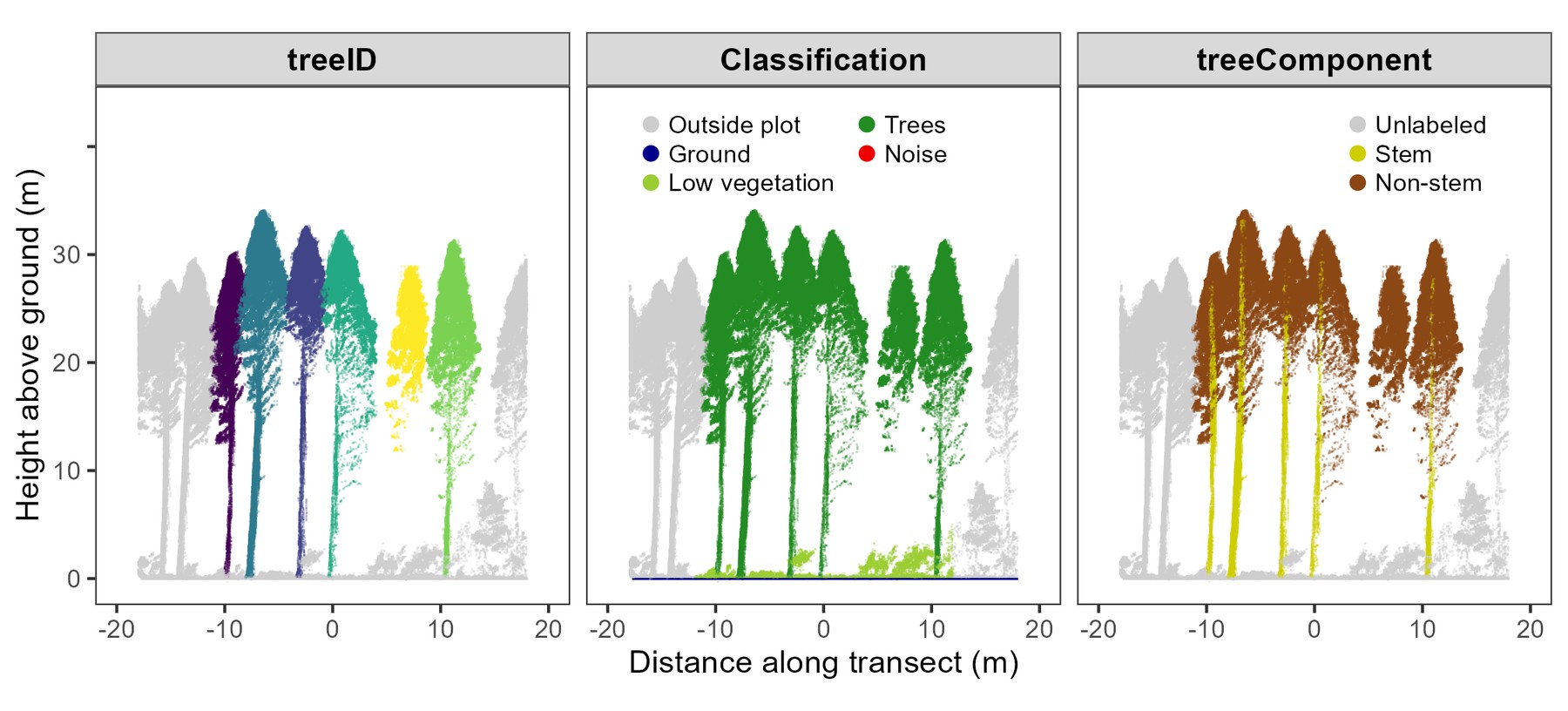}
  \caption{Example of point attributes for \emph{treeID}, \emph{Classification}, and \emph{treeComponent}. An example cross-section from Plot 5 at the Saiki site is presented.}
  \label{fig:6}
\end{figure}

\FloatBarrier
\section{Dataset validation}\label{dataset-validation}

\subsection{Validation}\label{validation}

To validate the manual annotation and tree correspondence between the LiDAR data and the field survey, tree heights derived from the LiDAR point clouds were compared with those from the field survey (Figure 7). Tree height from the LiDAR data was calculated as the height difference between the highest point of each annotated tree and the ground. For the UAV-LiDAR data, the root mean squared error (RMSE) of tree height was 1.24 m for Japanese cedar and 0.96 m for Japanese cypress after excluding dead trees. Furthermore, all field-surveyed trees were successfully matched to the annotated point clouds with a mean distance of 0.51 m between field survey tree locations and the centroid of the corresponding annotated point clouds. These results indicate good agreement between field survey data and annotated UAV-LiDAR data, supporting the accuracy of both manual annotation and tree matching. Although there were two growing seasons between the UAV-LiDAR acquisition and field survey measurements at the Saiki site, the resulting bias remained small. For the terrestrial LiDAR data, a larger bias (-2.94 m) and RMSE (3.85 m) were observed, which were expected because of canopy occlusion and the LiDAR sensor used in this study.

\begin{figure}[htbp]
  \centering
  \includegraphics[width=0.89\linewidth]{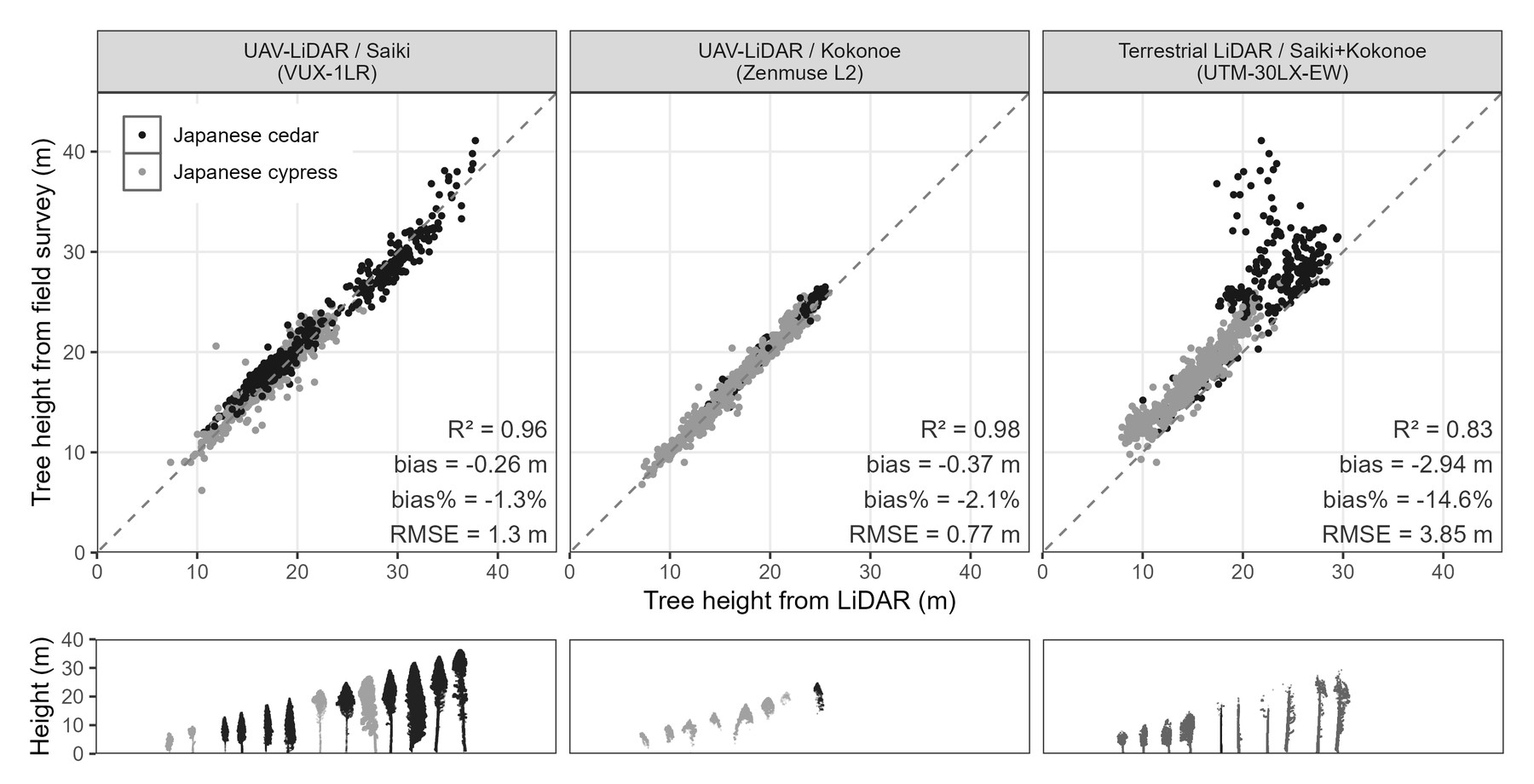}
  \caption{Comparison of tree heights measured in the field survey with those derived from the UAV-LiDAR and terrestrial LiDAR data (upper panels). Examples of annotated tree point clouds are shown in the lower panels.}
  \label{fig:7}
\end{figure}

\subsection{Uncertainty}\label{uncertainty}

Despite careful manual annotation, some uncertainty remains as branches or leaves of adjacent trees often overlap, making it difficult to separate individual trees. This issue occurs more frequently in Japanese cypress forests. In addition, suppressed trees are inherently more difficult to annotate because of partial or complete occlusion by dominant trees. For semantic labeling (stem and non-stem), exact separation of stem points from branches and leaves was challenging; thus, this classification may contain uncertainties. These uncertainties in the manual annotation process should be considered when conducting analyses.

In plot 13 at the Saiki site, several trees within the plot boundary were identified in the UAV-LiDAR point clouds; however, corresponding records in the field survey data were lacking. These discrepancies were likely caused by omissions during the field survey. To address this issue, new tree IDs were assigned to these trees; consequently, some \emph{treeID} in the point cloud data had no corresponding trees in the field survey data. For completeness, tree height and DBH for these trees were manually measured from the UAV-LiDAR point cloud and are provided in a separate file rather than within the field survey datasheet.

Several tree attributes in the field survey data involve inherent uncertainties. Although stem volume is provided for each tree, it was calculated using species specific stem volume equations. Consequently, stem volumes of individual trees may contain uncertainties. Previous studies have reported RMSEs of 2.5--13.5\% between harvested and calculated stem volumes for Japanese cedar and Japanese cypress trees \citep{inoueTheoreticalDerivationTwoway2001,mitsudaExaminationFactorsAffecting2012}. Additionally, stand age was obtained from forest management records provided by the Forestry Agency \citep{forestryagencyNationalForestResource2026}. These records may not accurately represent the actual stand conditions because of potential mismatches between historical management records and the current status of the stands.

The terrestrial LiDAR system used in this study is less suitable for measuring tree height, because of the characteristics of the laser sensor. This limitation is evident from the visual comparison of the point clouds (Figure 8) and the validation results presented above. Tree height measurements derived from these data should be interpreted with caution and are not recommended as a reference source. In contrast, terrestrial LiDAR data can generally provide accurate DBH measurements, and previous studies that used this terrestrial LiDAR sensor in Japanese planted forests have confirmed this, reporting RMSEs of 1.1--3.2 cm \citep{murokiEstimatingStandVolume2019,matsumuraAccuracyValidationVarious2020,suematsuInfluenceSamplingGrid2020,shimizuIntegratingTerrestrialLaser2022}. However, noise points were observed in some plots, and users should be aware of their presence and potential influence when analyzing point cloud data.

\begin{figure}[htbp]
  \centering
  \includegraphics[width=0.85\linewidth]{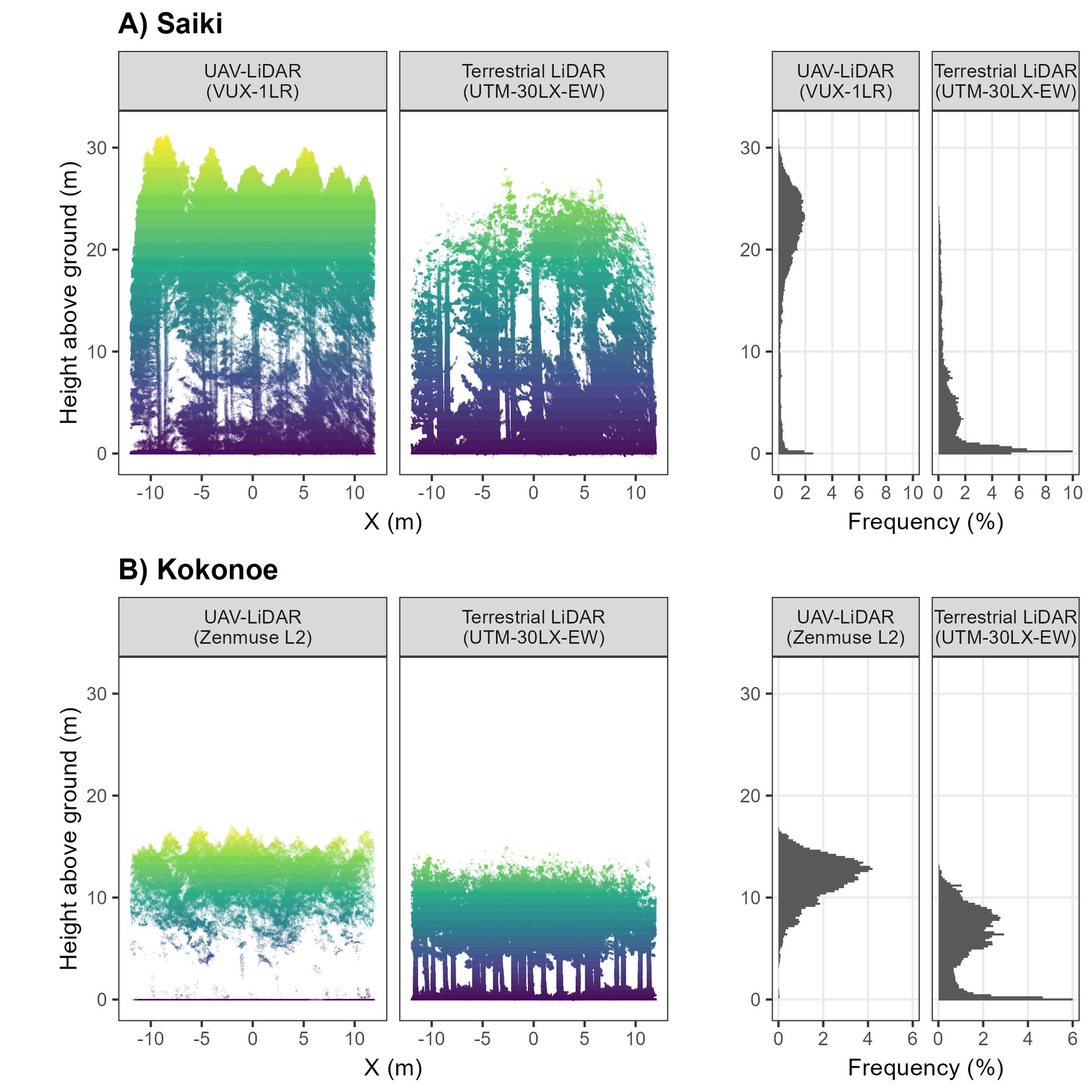}
  \caption{Example of normalized UAV-LiDAR and terrestrial LiDAR point clouds for (A) a Japanese cedar plot (Plot 1) in Saiki and (B) a Japanese cypress plot (Plot 1) in Kokonoe, shown in the left panels. The right panels show histograms of point clouds for each height bin after removing ground points.}
  \label{fig:8}
\end{figure}

\FloatBarrier
\section{Dataset usage}\label{dataset-usage}

CedarCypress3D is designed to provide ready-to-use, manually annotated UAV-LiDAR data for individual tree analysis, such as the development and validation of point cloud-based instance segmentation methods. For a subset of UAV-LiDAR data (22 out of 34 plots), semantic labels (stem and non-stem) are also available for individual trees; thus, it is suitable for semantic segmentation. However, this dataset can also be used for a wide range of other purposes. The field survey data provide tree-level reference measurements that could be used to develop and validate models for predicting tree attributes including DBH, tree height, and stem volume. In particular, the high point density of the UAV-LiDAR data at the Saiki site enables the direct estimation of stem diameter from the point clouds, offering opportunities to develop DBH estimation methods. Furthermore, the paired UAV-LiDAR, terrestrial LiDAR, and field survey data also facilitate comparative studies of different sensing platforms for individual tree analysis.

\FloatBarrier
\section*{Acknowledgments}\label{acknowledgments}

This study was supported by the Japan Forest Foundation through a Grant-in-Aid for Forestry Promotion and JSPS KAKENHI Grant Number JP24K01804. The authors thank the Forest Research and Management Organization for providing access to the UAV-LiDAR data. We are also grateful to Natsu Sannomiya for her assistance with the manual annotation of the LiDAR point cloud data. We thank Dr. Kazuo Hosoda and Dr. Gen Takao for their assistance with field data collection.

\FloatBarrier
\section*{Data availability}\label{data-availability}

The dataset is available at https://doi.org/10.5281/zenodo.22168721. The exact plot locations are available from the corresponding author upon reasonable request.

\FloatBarrier
\section*{Declaration of interest}\label{declaration-of-interest}

The authors report there are no competing interests to declare.

\FloatBarrier
\section*{Declaration of generative AI use}\label{declaration-of-generative-ai-use}

During the preparation of this work, the authors used ChatGPT and Claude to check grammatical errors and improve the language of the manuscript. After using these tools, the authors reviewed and edited the content as needed and take full responsibility for the content of the article.

\FloatBarrier
\section*{Author contribution}\label{author-contribution}

\textbf{Katsuto Shimizu:} Writing -- original draft, Writing -- review \& editing, Conceptualization, Data curation, Formal analysis, Methodology, Investigation, Resources, Validation. \textbf{Fumiaki Kitahara:} Writing -- review \& editing, Data curation, Formal analysis, Investigation. \textbf{Tomohiro Nishizono:} Methodology, Writing -- review \& editing, Data curation, Formal analysis, Investigation, Funding acquisition. \textbf{Hideki Saito:} Writing -- review \& editing, Investigation, Funding acquisition. \textbf{Masayoshi Takahashi:} Writing -- review \& editing, Investigation. \textbf{Shingo Obata:} Writing -- review \& editing, Investigation. \textbf{Shunsuke Tei:} Writing -- review \& editing, Investigation. \textbf{Naoyuki Furuya:} Writing -- review \& editing, Investigation, Data curation. \textbf{Tomoya Goto:} Writing -- review \& editing, Investigation. \textbf{Eiji Kodani:} Writing -- review \& editing, Investigation. \textbf{Yusuke Yamada:} Writing -- review \& editing, Investigation.

\bibliographystyle{apalike}
\bibliography{refs}

\end{document}